\documentclass[10pt,twocolumn,letterpaper]{article}

\usepackage{cvpr}              

\definecolor{cvprblue}{rgb}{0.21,0.49,0.74}
\usepackage[pagebackref,breaklinks,colorlinks,allcolors=cvprblue]{hyperref}
\usepackage{multirow}                 
\usepackage{multicol}                 
\usepackage{float}                    
\usepackage{makecell}                 
\usepackage{booktabs}                 
\usepackage{graphicx} 
\usepackage[dvipsnames]{xcolor}
\usepackage{colortbl}  
\usepackage{pifont}
\usepackage{amssymb}
\usepackage{xcolor}
\usepackage{marvosym}
\renewcommand{\thefootnote}{}

\def\paperID{4235} 
\def\confName{CVPR}
\def\confYear{2025}

\title{Evolving High-Quality Rendering and Reconstruction in a Unified Framework with Contribution-Adaptive Regularization
}

\author{You Shen\textsuperscript{1}\footnotemark[2]  , Zhipeng Zhang\footnotemark[2]  , Xinyang Li\textsuperscript{1}, Yansong Qu\textsuperscript{1}, Yu Lin\textsuperscript{1}, Shengchuan Zhang\textsuperscript{1}, Liujuan Cao\textsuperscript{1}\footnotemark[1] \\
\textsuperscript{1}Key Laboratory of Multimedia Trusted Perception and Efficient Computing,\\
Ministry of Education of China, Xiamen University
}

\begin{document}
\maketitle
\renewcommand{\thefootnote}{\fnsymbol{footnote}}
\footnotetext[1]{Corresponding Author.}
\footnotetext[2]{Equal Contribution.}

\begin{abstract}

Representing 3D scenes from multiview images is a core challenge in computer vision and graphics, which requires both precise rendering and accurate reconstruction. Recently, 3D Gaussian Splatting (3DGS) has garnered significant attention for its high-quality rendering and fast inference speed. Yet, due to the unstructured and irregular nature of Gaussian point clouds, ensuring accurate geometry reconstruction remains difficult. Existing methods primarily focus on geometry regularization, with common approaches including primitive-based and dual-model frameworks. However, the former suffers from inherent conflicts between rendering and reconstruction, while the latter is computationally and storage-intensive. To address these challenges, we propose CarGS, a unified model leveraging \textbf{C}ontribution-\textbf{a}daptive \textbf{r}egularization to achieve simultaneous, high-quality rendering and surface reconstruction. The essence of our framework is learning adaptive contribution for Gaussian primitives by squeezing the knowledge from geometry regularization into a compact MLP. Additionally, we introduce a geometry-guided densification strategy with clues from both normals and Signed Distance Fields (SDF) to improve the capability of capturing high-frequency details. Our design improves the mutual learning of the two tasks, meanwhile its unified structure doesn't require separate models as in dual-model based approaches, guaranteeing efficiency. Extensive experiments demonstrate CarGS’s ability to achieve state-of-the-art (SOTA) results in both rendering fidelity and reconstruction accuracy while maintaining real-time speed and minimal storage size. 

\end{abstract}    
\section{Introduction}
\label{sec:intro}

Novel view synthesis (NVS) and surface reconstruction from multi-view images are both challenging and critical tasks in the fields of computer vision and graphics. Both academia and industry are actively pursuing unified scene representations that enable high-quality rendering and accurate geometry reconstruction concurrently to address the growing demands of applications in areas such as augmented reality (AR), virtual reality (VR), robotics, and autonomous driving.
\begin{figure}[t]
	\centering
	\includegraphics[scale=0.46]{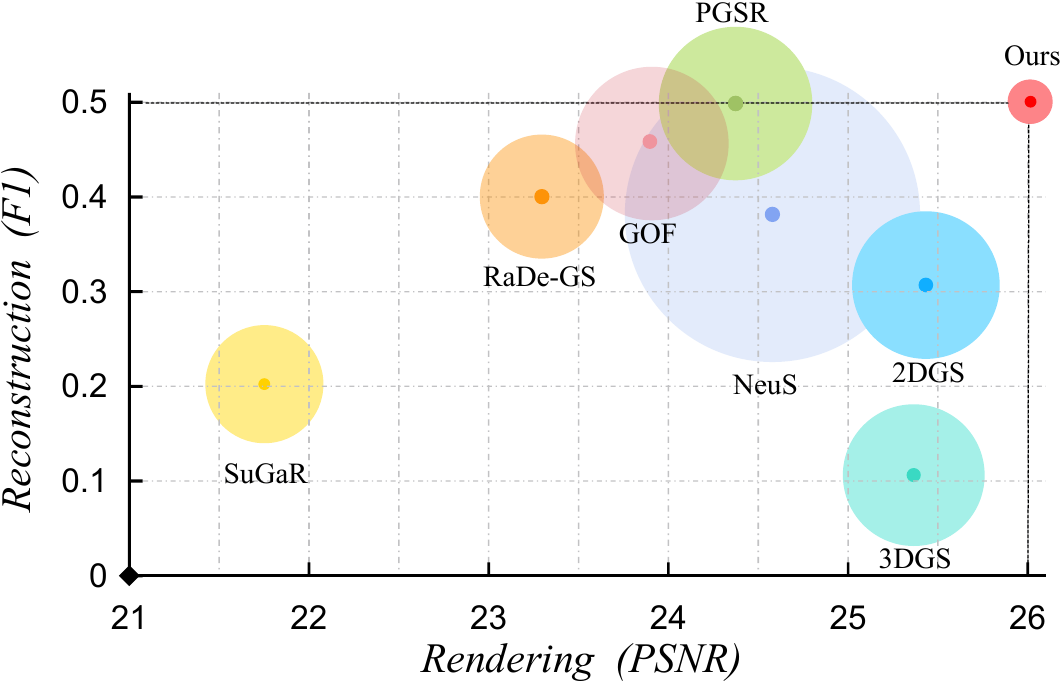}
	\caption{The performance of CarGS in TnT datasets, achieving high-quality rendering and reconstruction in a unified model. The storage size of the model is represented by the radius of the circle.}\label{P1}\vspace{-3mm}
\end{figure}
However, simultaneously achieving SOTA performance in both rendering and reconstruction with a unified model is non-trivial, as an inherent conflict exists between the two tasks~\cite{huang20242d,guedon2024sugar,chen2024pgsr}.


Revisiting recent approaches, we identify two solutions aimed at mitigating the conflict. The first branch, which we refer to as “primitive-based”, introduces additional geometric regularization to 3D Gaussian primitives to improve surface reconstruction. For example, Sugar~\cite{guedon2024sugar} regularizes 3D Gaussians to align more closely with surfaces. Although this approach ensures parameter efficiency, it leads to a misaligned prioritization between appearance and geometry, eventually degrading rendering performance. The second branch, termed “dual-model based”, seeks to guide the learning of one task through information derived from the other. For instance, GSDF~\cite{yu2024gsdf} introduces a dual-model architecture that jointly supervises the 3D Gaussian Splatting model for rendering and the SDF model for reconstruction. However, this strategy necessitates separate models for training and inference, resulting in much larger computational and storage costs. Consequently, a natural question arises: \textit{Is it possible to develop an efficient framework (\textbf{fast speed with small storage}) that is also effective \textbf{(SOTA performances on both tasks)}?}

\begin{figure}[!t]
	\centering
	\includegraphics[scale=0.44]{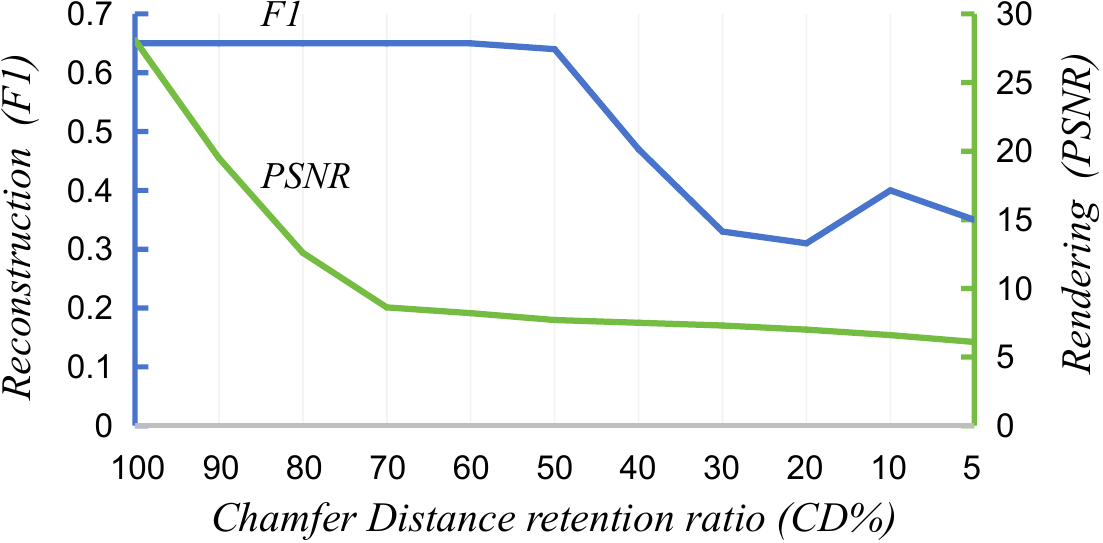}
	\caption{The performance of the Baseline model~\cite{chen2024pgsr} across various CD\% thresholds demonstrates that each Gaussian primitive contributes differently to specific tasks.}\label{P2}\vspace{-3mm}
\end{figure}

Before answering this question, we first present an intriguing observation that motivates us to pursue the ``unified goal'' from the perspective of learning adaptive contribution of Gaussian primitives. Specifically, with the baseline model PGSR~\cite{chen2024pgsr} trained on the ``Barn" scene from the Tanks and Temples dataset~\cite{knapitsch2017tanks}, we first calculate the Chamfer Distance (CD) from the center of each Gaussian to the ground truth surface. Then, as shown in Fig.~\ref{P2}, when retaining 50\% of the Gaussian points by discarding those with larger CD values, we find that reconstruction quality (F1) is surprisingly maintained, while rendering quality (PSNR) experiences a sharp decline. Conversely, when additional Gaussian points with smaller CDs are dropped, reconstruction quality begins to deteriorate, while rendering quality remains relatively stable. This observation suggests that, 1) despite geometric regularization encouraging Gaussian primitives to closely adhere to the surface~\cite{chen2024pgsr, huang20242d}, a significant number of floaters persist in the scene. 2) These floaters contribute less to reconstruction but are essential for rendering quality. More precisely, each Gaussian primitive contributes differently to each task.

Therefore, an intuitive idea is to build a unified model with adaptive contributions for each Gaussian primitive. This ``unified'' structure contrasts with dual-model designs, significantly reducing storage and computational demands from the outset. The ``adaptiveness'' mitigates conflicts in Gaussian primitive contributions when jointly optimizing for rendering and reconstruction tasks. To this end, we propose CarGS: a unified Gaussian Splatting model based on \textbf{C}ontribution-\textbf{a}daptive \textbf{r}egularization. Firstly, we squeeze geometry regularization knowledge into a lightweight residual structure to adaptively adjust each Gaussian primitive’s contribution for rendering and reconstruction. Then, to further capture geometric details, we introduce a geometry-guided densification strategy that controls Gaussian expansion using clues from both normals and SDF. Experimental results show that our method achieves state-of-the-art reconstruction performance within a unified model, without compromising rendering quality. Remarkably, our model consumes only 9\% of the storage and 40\% of the training time required by GSDF~\cite{yu2024gsdf} (dual-model based) while running 100× faster. Moreover, its rendering and reconstruction performance surpasses both GSDF and PGSR~\cite{chen2024pgsr} (primitive-based). We believe our approach establishes a strong baseline for unified rendering and reconstruction.

In summary, our main contributions are as follows: 1) We identify the key bottleneck in developing a unified, efficient, and effective framework for both rendering and reconstruction. 2) We propose to learn the adaptive contribution of each Gaussian primitive, addressing the inherent conflict in creating a unified framework for rendering and reconstruction. 3) We introduce a geometry-guided densification strategy to enhance the model's ability to capture fine details. 4) Our unified framework simultaneously achieves high-quality rendering and reconstruction while consuming significantly less storage than previous methods (see Fig.~\ref{P1}).

\section{Related work}
\label{sec:formatting}
\subsection{Novel View Synthesis}

Novel View Synthesis aims to generate new perspectives of a scene using a limited collection of existing views.


\noindent \textbf{NVS based on Neural Radiance Fields.} A pivotal advancement in this field was the introduction of Neural Radiance Fields (NeRF)~\cite{mildenhall2021nerf}. NeRF employs MLPs to map 3D spatial locations along rays to color and density values, which are subsequently aggregated through neural volumetric rendering to produce high-quality images. Subsequent research has focused on enhancing the capabilities of NeRF models through feature-grid representations~\cite{sun2022direct,fridovich2022plenoxels,kulhanek2023tetra} and improving rendering speed using baking techniques~\cite{hedman2021baking,reiser2021kilonerf,yariv2023bakedsdf}. Recently, NeRF has been adapted to address challenges such as anti-aliasing~\cite{barron2021mip} and the modeling of unbounded scenes~\cite{barron2022mip,zhang2020nerf++}.



\noindent \textbf{NVS based on 3D Gaussian Splatting.} Recently, 3D Gaussian Splatting~\cite{kerbl20233d} has emerged as a transformative approach in neural rendering by utilizing anisotropic 3D Gaussians as primitives. This method sorts Gaussians by depth and rasterizes them onto a 2D screen through alpha blending, resulting in impressive rendering quality and real-time speed. Subsequent research has enhanced 3D Gaussian Splatting by improving rendering quality~\cite{yu2024mip,cheng2024gaussianpro} and expanding its applicability to large-scale scenes~\cite{lu2024scaffold,zhang2024gaussian}. For example, Mip-splatting~\cite{yu2024mip} introduces anti-aliasing techniques to mitigate pixelation and blurring in high-resolution renders, thereby enhancing visual sharpness and consistency across viewpoints. Scaffold-GS~\cite{lu2024scaffold} employs anchor points to manage local 3D Gaussians and dynamically predicts their attributes based on viewing direction and distance, thereby optimizing the adaptive rendering process.

While these NVS methods excel in view synthesis, they inherently overlook the accurate reconstruction of scene geometry, resulting in blurred volumetric density fields that impede the extraction of high-quality surfaces.



\subsection{Neural Surface Reconstruction}
The success of neural rendering has sparked significant interest in neural surface reconstruction~\cite{wang2021neus, li2023neuralangelo, oechsle2021unisurf,wu2022voxurf,fu2022geo}. By leveraging coordinate-based neural networks, these methods encode scene geometry using occupancy fields or SDF values, facilitating a smooth and differentiable representation of 3D structures. While NeRF-based approaches, when combined with volume rendering, can produce smooth and complete surfaces, they often encounter difficulties in capturing fine details and exhibit slow optimization speeds. To overcome these limitations, recent advancements~\cite{li2023neuralangelo} have integrated multi-resolution hashed feature grids derived from Instant Neural Graphics Primitives~\cite{muller2022instant}, significantly enhancing representational capabilities and achieving SOTA results. Nonetheless, efficient reconstruction of large-scale scenes remains a challenge. For instance, Neuralangelo~\cite{li2023neuralangelo} requires 128 GPU hours to fully reconstruct a single scene from the Tanks and Temples dataset~\cite{knapitsch2017tanks}.

Motivated by the impressive efficiency of 3DGS in NVS, recent research has investigated the use of anisotropic 3D Gaussians as foundational primitives for scene geometry representation. SuGaR~\cite{sun2022direct} aligns 3D Gaussians with potential surfaces by approximating them as 2D planar primitives with binary opacity. 2D Gaussian Splatting~\cite{huang20242d} replaces 3D Gaussians with 2D counterparts, resulting in more accurate ray-splat intersections and regularizing depth maps to achieve a denser distribution of Gaussian primitives. PGSR~\cite{chen2024pgsr} employs both single-view and multi-view geometric regularization to attain globally consistent geometry. Despite these advancements in neural surface reconstruction, a fidelity gap persists, as the explicit regularization of Gaussian primitives constrains rendering quality. To address this limitation, GSDF~\cite{yu2024gsdf} introduces a dual-model framework that separates rendering (3DGS) and geometry (SDF) tasks, utilizing joint supervision during training to enhance both reconstruction and rendering. However, this design incurs additional training and storage costs due to the requirement of maintaining two independent models. Moreover, the geometry reconstruction branch is not capable of real-time inference.

Different from existing approaches, our paper presents a unified model designed to simultaneously achieve real-time and robust rendering and reconstruction tasks without compromising geometry accuracy and rendering quality.


\section{Method}
\label{sec:Method}
This section details the proposed framework. Particularly, Sec.~\ref{pre} presents preliminary studies on 3D Gaussian Splatting and its representative anchor-based paradigm. Then in Sec.~\ref{conflict}, we analyze the problem of contribution conflicts to identify the most significant attributes for geometry reconstruction. Sec.~\ref{litegeo} introduces Lite-Geo, a lightweight residual module that adapts the geometry contribution of Gaussian primitives. Based on Lite-Geo, we further propose a geometry-guided densification strategy in Sec.~\ref{densify} that leverages geometric cues to align Gaussians with the surface, enhancing both structural fidelity and representational capacity. Finally, in Sec.~\ref{optimization}, we introduce geometric regularization to guide the optimization of our modules, which improves the quality of geometry reconstruction.

\subsection{Preliminary}\label{pre}
\noindent \textbf{3D Gaussian Splatting.} 
3DGS models the scene with a set of explicit Gaussians that inherit the differentiable properties of volumetric representations, allowing efficient rendering via tile-based rasterization. 
Initially, each point generated by the Structure-from-Motion (SfM) serves as the position (mean) $\mu$ of a corresponding 3D Gaussian:
\begin{equation}
G(x) = e^{-\frac{1}{2} (x - \mu)^T \Sigma^{-1} (x - \mu)},
\label{eq:gaussian}
\end{equation}
where $x$ denotes an arbitrary position in the 3D scene, while $\Sigma$ represents the covariance matrix of the 3D Gaussian. $\Sigma$ is constructed using a scaling matrix $S$ and a rotation matrix $R$ to maintain positive semi-definiteness.
\begin{equation}
\Sigma = R S S^T R^T,\label{e2}
\end{equation}
Each 3D Gaussian is assigned with an opacity value $\alpha$, which is multiplied with $G(x)$ to adjust its contribution weight to rendering and reconstruction during the blending process. 3DGS renders scenes efficiently through tile-based rasterization instead of traditional ray marching. Each 3D Gaussian $G(x)$ is projected as a 2D Gaussian onto the image plane, after which a tile-based rasterizer sorts and blends the 2D Gaussians using $\alpha$-blending:
\begin{equation}
F(x') = \sum_{i \in N} f_i \sigma_i \prod_{j=1}^{i-1} (1 - \sigma_j), \quad \sigma_i = \alpha_i G_i(x)\label{equation_alpha}
\end{equation}
where $x'$ is the queried pixel location in the image plane and $N$ denotes the number of sorted 2D Gaussians associated with the queried pixel. For a Gaussian with 3D center $(x_i, y_i, z_i)$ and color $c_i$, $f_i$ is assigned to $c_i$ in the rendering task, and to $z_i$ for depth blending in the reconstruction task.


\noindent \textbf{Anchor-based Gasussian Splatting.}
Our goal is to simultaneously achieve high-quality rendering and reconstruction while maintaining efficiency. In our work, we adopt anchor-based GS modeling as our baseline due to its balanced efficiency and effectiveness. A representative work in this paradigm, Scaffold-GS~\cite{lu2024scaffold}, enhances original 3DGS by improving scene structure fidelity and incorporating view clues. Specifically, it constructs a hierarchical 3D Gaussian framework with anchor points that encode local scene details and generate local neural Gaussians,
\begin{equation}
\{\mu_0, \dots, \mu_{k-1}\} = x + \{{o}_0, \dots, {o}_{k-1}\} \cdot \gamma,
\end{equation}
where $o_{k} \in \mathbb{R}^{3}$ denotes the learned offsets, and $\gamma$ represents the scaling factor. For each anchor, the optimized attributes (color, center, variance, and opacity) of the $k$ neural Gaussians are decoded through an MLP, $M$, based on the anchor feature $\mathbf{f}$, as well as the distance $\delta$ and direction $\theta$ from the camera to the anchor point. Specifically, the opacity values for neural Gaussians generated from an anchor point are,
\begin{equation}
{\alpha_0, \dots, \alpha_{k-1}} = M(\mathbf{f},  \delta, 
 \theta),
\end{equation}
\begin{figure}[t]
	\centering
	\includegraphics[scale=0.52]{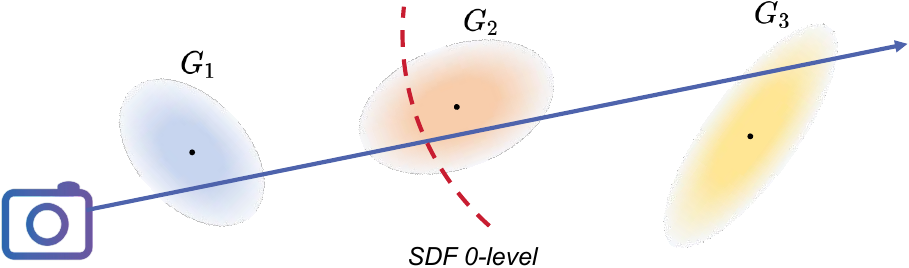}
	\caption{The illustration of contribution conflict between rendering and reconstruction tasks.}\label{P_3}\vspace{-3mm}
\end{figure}
\subsection{Analysis}
\label{sec:analysis}
In this section, we analyze the underlying reasons for the rendering performance degradation when geometry regularization is applied to 3DGS primitives. Then, we conduct experiments to analyze the effect of Gaussian's internal attributes to geometry reconstruction. 

\noindent\textbf{Analysis of Contribution Conflicts.}\label{conflict} In Fig~\ref{P_3}, we consider a simple demonstration of alpha-blending along a ray intersecting three Gaussian primitives. As shown, $G_{2}$ is close to the real surface, $G_{1}$ and $G_{3}$ are farther away. According to Eq.~\ref{equation_alpha}, the color $C(x')$ and depth $D(x')$ corresponding to the queried pixel along the ray is calculated as follows:
\begin{equation}
C(x')=c_{1}w_{1}+c_{2}w_{2}+c_{3}w_{3},
\end{equation}
\begin{equation}
D(x')=z_{1}w_{1}+z_{2}w_{2}+z_{3}w_{3}
\end{equation}
\begin{equation}
w_{i}=\sigma_i \prod_{j=1}^{i-1} (1 - \sigma_j),\quad \sigma_i = \alpha_i G_i(x)\label{e8}
\end{equation}
where  $w_i$ represents the contribution of the Gaussian and serves as the weighting factor for ray integration. When only applying rendering regularization, $w_{1}$ and $w_{3}$ tend to increase to capture richer textures, as proved in Fig.~\ref{P2}. Conversely, when geometry regularization is applied, the values of $w_{1}$ and $w_{3}$ tend to decrease due to $z_1$ and $z_3$ deviating from the SDF 0-level set, while $w_{2}$ increases to provide more consistent depth information. This shift in the contribution weights $w_{i}$ creates conflicts, which can lead to a decline in rendering performance when geometry regularization is simply added when training the rendering task~\cite{huang20242d,chen2024pgsr}.

\begin{table}[]
\tabcolsep=0.51cm
\begin{center}
\renewcommand{\arraystretch}{1.2}
\begin{tabular}{c|c|c|c}
\Xhline{1px}\hline
     & Cov & Opacity & Baseline \\ \hline
F1$\uparrow$   & 0.46       & 0.64           & 0.65     \\ \hline
PSNR$\uparrow$ & 23.84      & 28.72          & 28.76    \\ \hline
\end{tabular}
\caption{The importance of different attributes in Gaussian's contribution for the geometry reconstruction task.}\label{t1}
\end{center}\vspace{-4mm}
\end{table}

\noindent\textbf{Analysis of Attributes.} 
As discussed, the degradation in rendering performance after using geometry regularization stems from contribution conflicts. To further investigate this issue, we analyze which internal attributes of the Gaussian’s contribution are most susceptible to the effects of geometry regularization. As shown in Eq.~\ref{e8}, Gaussian's contribution $w_i$ is defined by its opacity $\alpha$ and Gaussian sampling value $G(x)$. Lu et al.~\cite{lu2024scaffold} emphasize that stable Gaussian positions are crucial for maintaining scene structure, suggesting that adjustments to $\mu$ in Eq.~\ref{eq:gaussian} remain relatively minor during optimization. Consequently, changes in $G(x)$ are more likely to arise from covariance in the covariance $\Sigma$ in Eq.~\ref{eq:gaussian}. 


Then, to investigate the roles of $\alpha$ (opacity) and $\Sigma$ (covariance) within the geometry regularization process, we conduct analysis experiments in Tab.~\ref{t1}. Specifically, with the baseline model PGSR~\cite{chen2024pgsr} on TnT's ``Barn'' scene, we respectively detach pacity and covariance from the geometry supervision to stop their gradigents
and then assess their independent effect on the performance of geometry reconstruction. As shown in Tab.~\ref{t1}, excluding the gradient of opacity from geometry optimization shows minimal impact on reconstruction quality, while excluding covariance's gradient severely impairs geometry learning. Hence, covariance plays the essential role in regulating the contributions of Gaussian primitives, making it important to tailor covariance for both high-quality rendering and reconstruction.

\begin{figure}[t]
	\centering
	\includegraphics[scale=0.55]{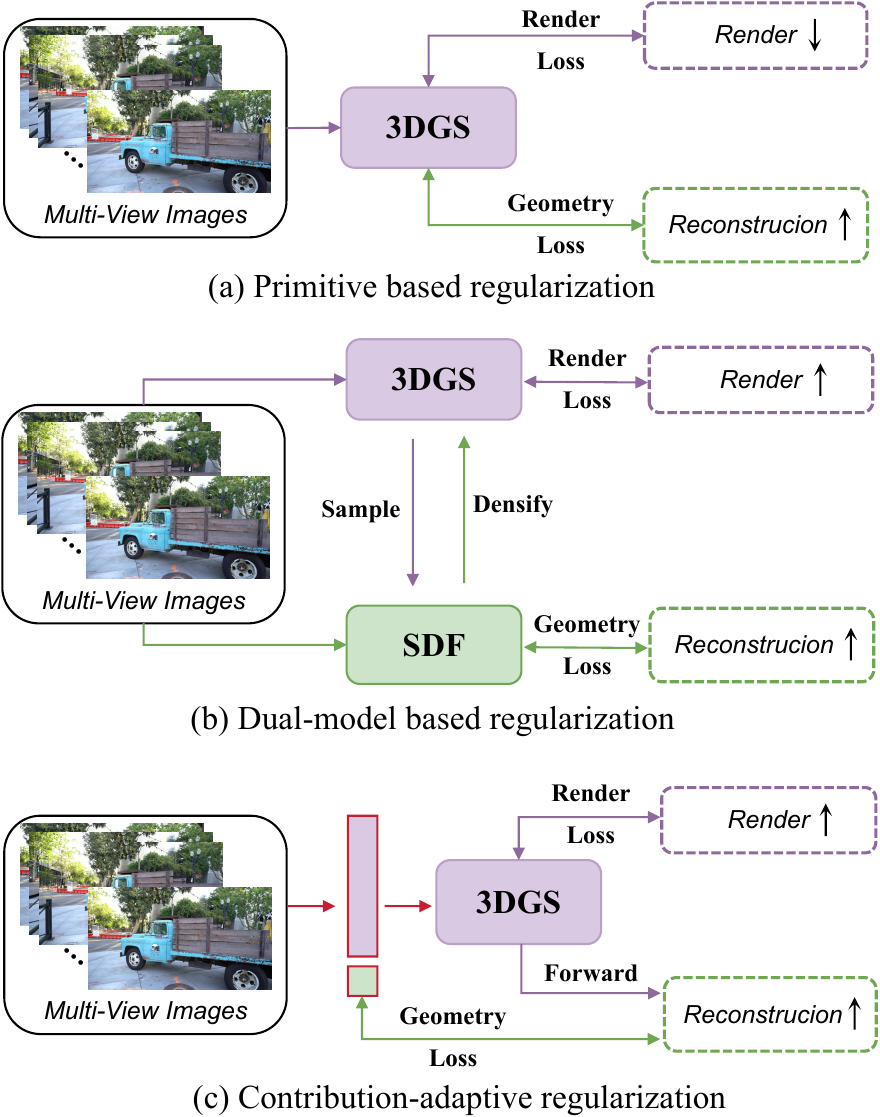}
	\caption{The illustration of the geometry regularization GS framework: (a) Primitive based regularization paradigm. (b) Dual-model-based regularization paradigm. (c) the proposed contribution adaptative regularization paradigm. }\label{P_5}\vspace{-3mm}
\end{figure}
\subsection{3DGS with Adaptive Contribution} \label{litegeo}
As shown in Fig.~\ref{P_5}, existing approaches for realizing rendering and reconstruction are categorized into primitive based~\cite{huang20242d,chen2024pgsr} and dual-model based~\cite{yu2024gsdf} frameworks. The former supervises Gaussian primitives with both rendering and geometry losses, leading to conflicting contributions and, consequently, performance degradation. In contrast, dual-model based methods train separate models for each task, \textit{i.e.,} a 3DGS model for rendering and an SDF model for reconstruction. While enhancing both tasks, it requires substantial computational and storage resources (see Tab.~\ref{t_eff}). Building on the above analysis, we are motivated to propose a unified and efficient contribution-adaptive 3DGS architecture to simultaneously achieve state-of-the-art performances by alleviating the adverse effects on rendering when using geometry regularization. 

\begin{figure*}[!t]
	\centering
	\includegraphics[scale=0.56]{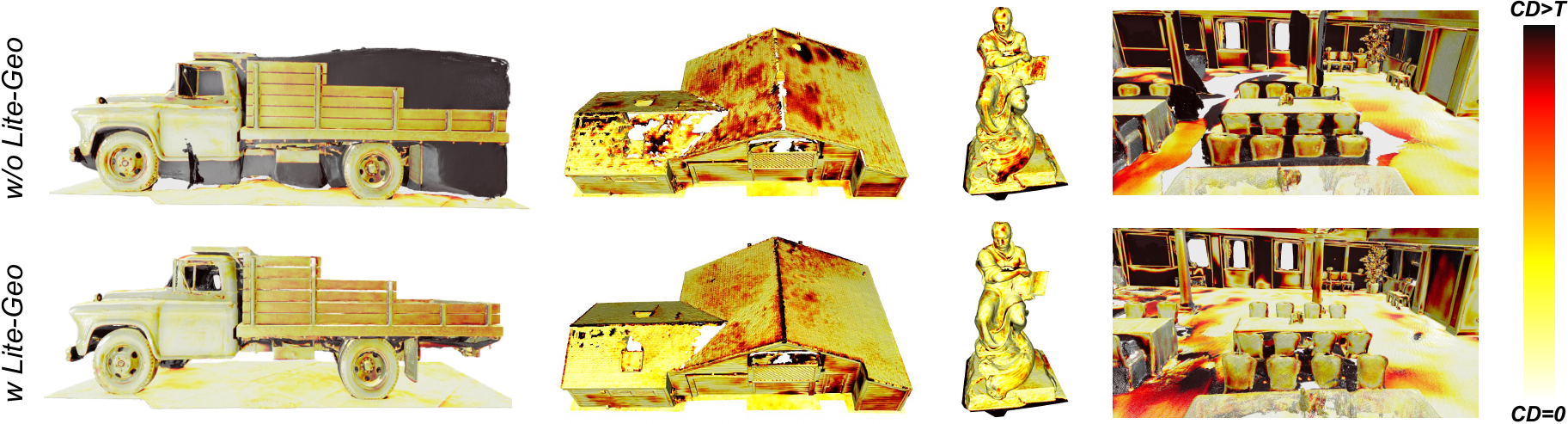}\vspace{-2mm}
	\caption{\textbf{The illustration of geometry overfitting.} If the Geo-MLP is solely associated with geometry reconstruction, notable overfitting is observed. By leveraging the approximate depth knowledge provided by rendering, Lite-Geo can effectively reduce overfitting and suppress noise where the Chamfer Distance exceeds the threshold.}\label{lite_geo}\vspace{-4mm}
\end{figure*}

\noindent\textbf{Geo-MLP.} In Sec.~\ref{sec:analysis}, we have identified covariance as the primary factor contributing to conflicts in geometry and rendering. Thus, we propose decoupling Gaussian contributions by predicting an additional covariance parameter dedicated to geometry reconstruction, while retaining the original covariance for rendering. As outlined in Eq.~\ref{e2}, covariance is decomposed into scaling and rotation matrices. Therefore, we first attempt to introduce an auxiliary MLP that takes the anchor feature $\mathbf{f}$ and view direction as inputs, predicting the scaling and rotation matrix. The predicted geometry-related scaling and rotation are supervised by reconstruction loss, as shown in Fig.~\ref{P_5} (c). However, 
nothing is as simple as it seems. As visualized in Fig.~\ref{lite_geo} (top), this straightforward design results in overfitting in geometry estimation. This challenge arises because conventional geometric regularization primarily enforces continuity and consistency, but lacks accurate supervision for depth, which is predominantly informed by rendering-based supervision.



\noindent \textbf{Lite-Geo.} To address this challenge, we introduce Lite-Geo, designed to inherit the Gaussian primitive’s contribution from the rendering task, which encodes inherent depth information after rendering optimization. This implicit depth information helps mitigate overfitting brought by the sole geometry reconstruction loss applied to Geo-MLP. More precisely, we structure the learning of covariance for geometry reconstruction as a residual update, where it is obtained by adding the offset predicted by Geo-MLP ($M_{\Sigma}^{geo}$) to the rendering covariance.



\begin{equation}
    y= M_{\Sigma}^{rgb} (\mathbf{f},\theta ;\phi _{1} )+\bigtriangleup y,
\end{equation}

\begin{equation}
\bigtriangleup y=M_{\Sigma}^{geo} (\mathbf{f},\theta;\phi _{2}),\phi _{2}=\lambda \phi _{1} 
\end{equation}
where $\phi _{1}$ and $\phi _{2}$  denote the parameters of $M_{\Sigma}^{rgb}$ and $M_{\Sigma}^{geo}$, respectively. $y$ and $\bigtriangleup y$ are 7-dimensional vectors, with the first three dimensions representing scale and the remaining four representing rotation through quaternions. This approach allows the depth information in rendering covariance to effectively inform the depth estimation in geometry reconstruction. To further reduce noise and enhance convergence, we initialize $\phi _{1}$ based on $\phi _{2}$ using a controllable scaling factor $\lambda$. Specifically, the rendering task is pre-trained independently for several epochs before incorporating the Lite-Geo module.



As shown in Fig.~\ref{lite_geo}, this residual structure effectively suppresses geometry overfitting. It also constrains outliers with high Chamfer Distance values that negatively impact rendering and reconstruction quality.


\begin{figure}[t]
	\centering
	\includegraphics[scale=0.6]{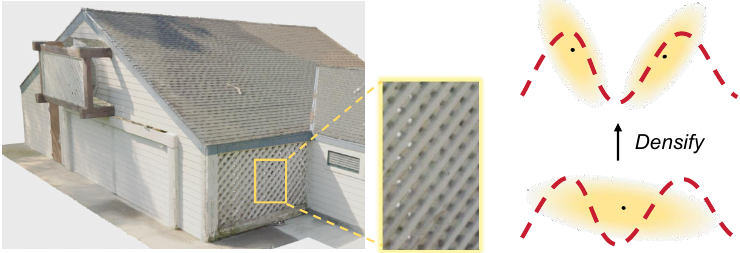}
	\caption{The illustration of high-frequency details, along with the proposed densification strategy which facilitates Gaussian growing with clues from both normals and SDF.}\label{P_4}\vspace{-3mm}
\end{figure}
\subsection{Geometry-guided Densification}\label{densify}
We further enhance the distribution of Gaussian primitives through a geometry-guided densification strategy. Our motivation stems from the observation that previous geometry-based densification methods~\cite{yu2024gsdf} rely solely on SDF values to control Gaussian growing, which limits their ability to capture fine details in high-frequency regions and often results in blurred rendering and overly smooth reconstructions. As shown in Fig.~\ref{P_4}, to better accommodate high-frequency details, we introduce normal vector clues alongside the original SDF information to guide Gaussian growing. Specifically, for a Gaussian anchor located at position $x$ and the SDF value within a certain range $\theta$, we evaluate its normal difference with the normal map by $n = N_d(x) - N(x)$. The SDF is obtained from the depth map by $s = D(x) - Z(x)$. Detailed calculations of normals and SDF are provided in the supplementary. We then define the criteria for Gaussian growing as follows:
\begin{equation}
\epsilon_g = \nabla_g + \omega_g \left( \zeta(s) \cdot (\omega_n n \text{ if } \mu(s) < \theta \text{ else } 1) \right),
\end{equation}
\begin{equation}
\zeta(s)=\exp \left( -\frac{s^2}{2\sigma^2} \right),
\end{equation}
where $\nabla_g$ is the averaged gradient of the Gaussian primitives accumulated over several training iterations. $\zeta(s)$ is a Gaussian function that transforms the SDF to a positive number, which monotonically decreases as the distance from the zero level set increases. The parameter $\omega_g$ controls the influence of geometric guidance. New Gaussian primitives are generated when $\epsilon_g$ exceeds a predefined threshold.

\subsection{Optimization}\label{optimization}
Relying solely on photometric loss to optimize 3DGS introduces noise into the output, as multi-view 3D reconstruction lacks sufficient constraints~\cite{zhang2020nerf++, barron2022mip} for geometric continuity. To improve the recovery of shape detail, we adopt the geometry regularization from PGSR~\cite{chen2024pgsr}, incorporating additional plane constraint and cross-view constraint. To ensure the paper is self-contained, we provide a detailed explanation of these two regularization terms.

\noindent \textbf{Plane Constraint.} To ensure local depth-normal consistency, each pixel and its neighbors should define a relatively smooth plane. Using the depths of a pixel and its neighboring pixels, we can generate a normal map, which is then optimized to align with the rendered normal map, thereby enforcing geometric consistency. The loss based on the plane constraint is defined as follows:
\begin{equation}
\mathcal{L}_{\text{plane}} = \frac{1}{|\mathcal{W}|} \sum_{x \in \mathcal{W}} \left\| N_d(x) - N(x) \right\|_1,
\end{equation}
$N_d(x)$ and $N(x)$ represent the normal derived from the depth gradient and the normal of the splats, respectively. $\mathcal{W}$ denotes the image pixels.

\noindent \textbf{Cross-view Constraint.} 
Plane constraint ensures local depth-normal consistency but lacks cross-view consistency. Hence, cross-view constraintsd should be introduced to improve global geometric consistency. Specifically, a pixel $x_r$ in the reference frame, associated with a normal $n_r$ and distance $d_r$, can be transformed to a corresponding pixel $x_n$ in a neighboring frame with the homography matrix $H_{rn}$. After forward projection from the reference frame to the neighboring frame, a backward projection $H_{nr}$ is applied. Then the consistency is ensured by minimizing the forward-backward projection error:
\begin{equation}
\mathcal{L}_{cross} = \frac{1}{|\mathcal{W}|} \sum_{x_r \in \mathcal{W}} \left\| x_r - H_{nr} H_{rn} x_r \right\|.
\end{equation}

\noindent \textbf{Final Loss.} Finally, we optimize our model by minimizing, 
\begin{equation}
\mathcal{L} = \mathcal{L}_c + \alpha \mathcal{L}_d + \beta \mathcal{L}_n
\label{e_18}
\end{equation}
where $\mathcal{L}_c$ is the RGB reconstruction loss that combines L1 loss with D-SSIM from~\cite{kerbl20233d}, while $\mathcal{L}_{\text{plane}}$ and $\mathcal{L}_{\text{cross}}$ represent the geometry regularization terms.

\begin{table*}[!t]
\small
\tabcolsep=0.11cm
\begin{center}
\renewcommand{\arraystretch}{1.4}
\begin{tabular}{ccccccccccccccc}
\Xhline{1px}\hline
            & \multicolumn{7}{c}{F1-Score$\uparrow$}                              &  & \multicolumn{6}{c}{PSNR$\uparrow$}                         \\ \cline{2-8} \cline{10-15} 
            & COLMAP & NeuS & Neurlangelo & 2DGS & GOF  & PGSR & CarGS &  & NeuS  & 2DGS  & RaDe-GS & GOF   & PGSR  & CarGS \\ \hline
Barn        & 0.55   & 0.29 & \textbf{\textcolor{OliveGreen}{0.70}}        & 0.36 & 0.51 & \textcolor{RoyalBlue}{0.66} & 0.60   &  & 26.36 & 28.79 & 26.45   & 26.71 & 28.80 & \textbf{\textcolor{OliveGreen}{28.91}}  \\
Caterpillar & 0.01   & 0.29 & 0.36        & 0.23 & \textcolor{RoyalBlue}{0.41} & \textcolor{RoyalBlue}{0.41} & \textbf{\textcolor{OliveGreen}{0.43}}   &  & \textbf{\textcolor{OliveGreen}{25.21}} & 24.23 & 22.45   & 22.90 & 23.04 & \textcolor{RoyalBlue}{25.16}  \\
Courthouse  & 0.11   & 0.17 & \textbf{\textcolor{OliveGreen}{0.28}}        & 0.13 & \textbf{\textcolor{OliveGreen}{0.28}} & 0.21 & 0.17   &  & \textbf{\textcolor{OliveGreen}{23.55}} & 23.51 & 19.73   & 21.44 & 21.81 & \textcolor{RoyalBlue}{23.54}  \\
Ignatius    & 0.22   & 0.83 & \textbf{\textcolor{OliveGreen}{0.89}}        & 0.44 & 0.68 & \textcolor{RoyalBlue}{0.80} & 0.78   &  & 23.27 & \textcolor{RoyalBlue}{23.82} & 21.57   & 22.20 & 22.21 & \textbf{\textcolor{OliveGreen}{24.77}}  \\
Meetingroom & 0.19   & 0.24 & \textcolor{RoyalBlue}{0.32}        & 0.16 & 0.28 & 0.29 & \textbf{\textcolor{OliveGreen}{0.35}}   &  & 25.38 & \textcolor{RoyalBlue}{26.15} & 25.52   & 25.72 & 24.72 & \textbf{\textcolor{OliveGreen}{27.27}}  \\
Truck       & 0.19   & 0.45 & 0.48        & 0.26 & 0.58 & \textcolor{RoyalBlue}{0.60} & \textbf{\textcolor{OliveGreen}{0.65}}   &  & 23.71 &\textbf{ \textcolor{OliveGreen}{26.85}} & 23.72   & 24.32 & 26.12 & \textcolor{RoyalBlue}{26.41}  \\ \hline
Mean        & 0.21   & 0.38 &\textbf{ \textcolor{OliveGreen}{0.50} }       & 0.30 & 0.46 & \textbf{\textcolor{OliveGreen}{0.50}} & \textbf{\textcolor{OliveGreen}{0.50}}   &  & 24.58 & \textcolor{RoyalBlue}{25.56} & 23.24   & 23.88 & 24.45 & \textbf{\textcolor{OliveGreen}{26.01}}  \\ \hline
\end{tabular}\vspace{-2mm}
\caption{\textbf{Quantitative result on Tanks and Temples dataset.} CarGS achieves the SOTA performance on surface reconstruction and image synthesis. The color \textcolor{OliveGreen}{green} indicates the best result and \textcolor{RoyalBlue}{blue} indicates the second best one.} 
\label{t2}
\end{center}\vspace{-4mm}
\end{table*}

\begin{table*}[!t]
\small
\tabcolsep=0.3cm
\begin{center}
\renewcommand{\arraystretch}{1.3}
\begin{tabular}{ccccc|ccc|ccc}
\Xhline{1px}\hline
 &  & \multicolumn{3}{c}{Indoor scenes} & \multicolumn{3}{c}{Outdoor scenes} & \multicolumn{3}{c}{Average on all scenes} \\
 &  & PSNR$\uparrow$ & SSIM$\uparrow$ & LPIPS$\downarrow$ & PSNR$\uparrow$ & SSIM$\uparrow$ & LPIPS$\downarrow$ & PSNR$\uparrow$ & SSIM$\uparrow$ & LPIPS$\downarrow$ \\ \hline
\multicolumn{1}{c|}{\multirow{4}{*}{\rotatebox{90}{Implicit}}} & \multicolumn{1}{c|}{NeRF} & 26.84 & 0.790 & 0.370 & 21.46 & 0.458 & 0.515 & 24.15 & 0.624 & 0.443 \\
\multicolumn{1}{c|}{} & \multicolumn{1}{c|}{Deep Blending} & 26.40 & 0.844 & 0.261 & 21.54 & 0.524 & 0.364 & 23.97 & 0.684 & 0.313 \\
\multicolumn{1}{c|}{} & \multicolumn{1}{c|}{INGP} & 29.15 & 0.880 & 0.216 & 22.90 & 0.566 & 0.371 & 26.03 & 0.723 & 0.294 \\
\multicolumn{1}{c|}{} & \multicolumn{1}{c|}{NeuS} & 25.10 & 0.789 & 0.319 & 21.93 & 0.629 & 0.600 & 23.74 & 0.720 & 0.439 \\ \hline
\multicolumn{1}{c|}{\multirow{5}{*}{\rotatebox{90}{Explicit}}} & \multicolumn{1}{c|}{3DGS} & \textbf{\textcolor{OliveGreen}{30.99}} & 0.926 & 0.199 & 24.24 & 0.705 & 0.283 & 27.24 & 0.803 & 0.246 \\
\multicolumn{1}{c|}{} & \multicolumn{1}{c|}{SuGaR} & 29.44 & 0.911 & 0.216 & 22.76 & 0.631 & 0.349 & 26.10 & 0.771 & 0.283 \\
\multicolumn{1}{c|}{} & \multicolumn{1}{c|}{2DGS} & 30.39 & 0.923 & 0.183 & 24.33 & 0.709 & 0.284 & 27.03 & 0.804 & 0.239 \\
\multicolumn{1}{c|}{} & \multicolumn{1}{c|}{PGSR} & 30.41 & \textbf{\textcolor{OliveGreen}{0.928}} & \textbf{\textcolor{OliveGreen}{0.167}} & \textcolor{RoyalBlue}{24.45} & \textcolor{RoyalBlue}{0.730} & \textbf{\textcolor{OliveGreen}{0.224}} & \textcolor{RoyalBlue}{27.43} & \textcolor{RoyalBlue}{0.830} & \textbf{\textcolor{OliveGreen}{0.193}} \\
\multicolumn{1}{c|}{} & \multicolumn{1}{c|}{CarGS} & \textcolor{RoyalBlue}{30.85} & \textbf{\textcolor{OliveGreen}{0.928}} & \textcolor{RoyalBlue}{0.171} & \textbf{\textcolor{OliveGreen}{24.51}} &\textbf{ \textcolor{OliveGreen}{0.741}} & \textcolor{RoyalBlue}{0.246} & \textbf{\textcolor{OliveGreen}{27.68}} &\textbf{ \textcolor{OliveGreen}{0.834}} & \textcolor{RoyalBlue}{0.209} \\ \hline
\end{tabular}\vspace{-2mm}
\caption{\textbf{Quantitative result of rendering quality on Mip-NeRF360 dataset.} The proposed CarGS achieves the best average performance on novel view synthesis compared with the representative implicit Nerf-based and explicit GS-based methods.}\vspace{-7.5mm}
\label{tab3}
\end{center}
\end{table*}

\section{Experiments}
\subsection{Experimental Setup}
\noindent \textbf{Datasets.} To validate the effectiveness of our method, we conduct experiments on diverse real-world datasets, including both indoor and outdoor environments. With the camera poses provided by these datasets, we employ COLMAP~\cite{schonberger2016structure} to generate an initial sparse point cloud for each scene. We select large and complex scenes from the TnT dataset~\cite{knapitsch2017tanks} to evaluate the quality of surface reconstruction. The widely used Mip-NeRF360 dataset~\cite{barron2022mip} is used to assess novel view synthesis performance.

\noindent \textbf{Evaluation protocols.} For fair comparisons with previous methods, we select three widely used image evaluation metrics to validate novel view synthesis: peak signal-to-noise ratio (PSNR), structural similarity index measure (SSIM), and learned perceptual image patch similarity (LPIPS)~\cite{zhang2018unreasonable}. We employ the F1-score to assess the surface quality on the TnT dataset. Apart from the commonly used metrics, we additionally report the storage size (MB) for model compactness and the rendering speed (FPS)  for efficiency.

\begin{figure*}[ht]
	\centering
	\includegraphics[scale=0.8]{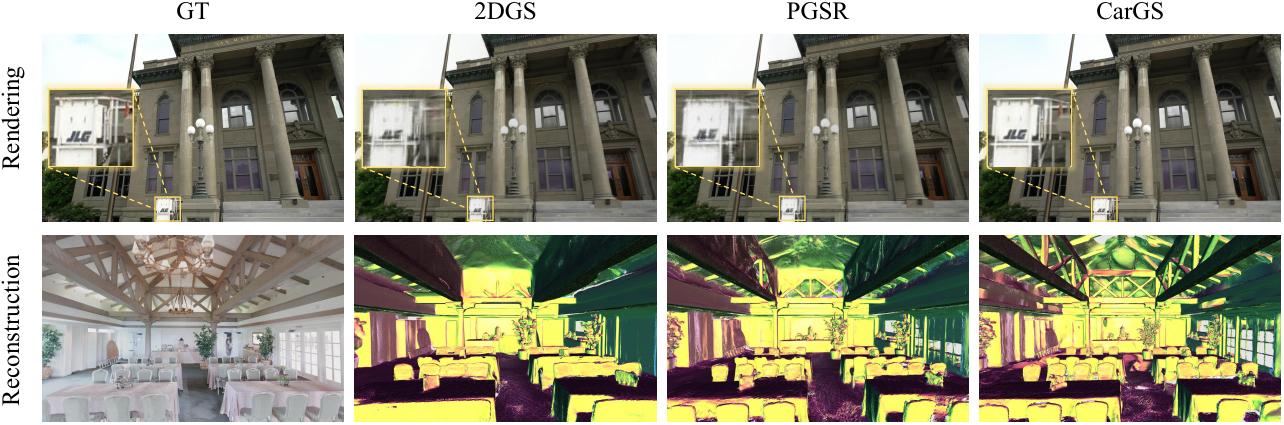}
	\caption{\textbf{Qualitative comparison on Tanks and Temples dataset. } In terms of rendering, our approach surpasses the baseline methods in effectively suppressing blurry regions. For reconstruction, we assess surface quality by visualizing the normal map generated from the reconstructed mesh. CarGS demonstrates advancements in capturing details of the scene compared to other approaches.}\label{P_qua}\vspace{-1mm}
\end{figure*}

\noindent \textbf{Implementation Details.}
Most of our training strategies and hyper-parameters are consistent with those used in 3DGS~\cite{kerbl20233d}. Specifically, we adopt a similar approach to halt densification at 15,000 iterations and optimize all models for 30,000 iterations. All MLPs in our method are 2-layer networks with ReLU activation function, and each hidden layer contains 32 units. The loss weights  $\alpha$ and $\beta$ in Eq.~\ref{e_18} are set to 0.01 and 0.2, respectively. We begin by rendering depth for each training view, then apply the TSDF Fusion algorithm~\cite{newcombe2011kinectfusion} to create the corresponding TSDF field, from which we subsequently extract the mesh~\cite{lorensen1998marching}. All experiments were conducted on an Nvidia RTX 3090 GPU.

\subsection{Result Analysis}
\noindent \textbf{Rendering Comparison.} As shown in Tab.~\ref{t2} and Tab.~\ref{tab3}, our method achieves rendering quality comparable to recent state-of-the-art 3DGS and NeRF-based methods. Unlike prior primitive-based approaches that apply shared Gaussian primitives for both rendering and reconstruction (e.g., 2DGS~\cite{huang20242d}, PGSR~\cite{chen2024pgsr}), our approach adaptively adjusts contributions to meet specific task requirements, allowing high-quality geometry reconstruction without sacrificing rendering fidelity. Our method also diverges from dual-model approaches (e.g., GSDF~\cite{yu2024gsdf}) by supporting joint training within a unified network for both tasks. As shown in Fig.~\ref{P_qua}, our approach shows better quality in rendering texture-less regions, where conventional 3D Gaussians struggle due to low accumulated gradients. 


\noindent \textbf{Reconstruction Comparison.} We compare our method, CarGS, with current state-of-the-art neural surface reconstruction approaches, including NeuS~\cite{wang2021neus} and NeuralAngelo~\cite{li2023neuralangelo}, as well as recent 3DGS-based methods like 2DGS~\cite{huang20242d}, GOF~\cite{yu2024gaussian}, and PGSR~\cite{chen2024pgsr}. Reconstruction performance is quantitatively evaluated on the TNT dataset, where CarGS achieves an F1 score comparable to NeuralAngelo and PGSR, outperforming other existing methods, as shown in Tab.\ref{t2}. Fig.\ref{P_qua} illustrates visual results from various Gaussian splatting approaches, with CarGS generating smoother and more precise shapes.



\begin{table}[]
\small
\tabcolsep=0.15cm
\begin{center}
\renewcommand{\arraystretch}{1.5}
\begin{tabular}{c|cccccc}
\Xhline{1px}\hline
 & PSNR$\uparrow$ & F1$\uparrow$ & Time$\downarrow$ & Storage$\downarrow$ & FPS$\uparrow$ \\ \hline
NeuS & 23.71 & 0.45 & \textgreater{}24h & 1.4GB & \textless{}1 \\
Neurlangelo & 25.43 & 0.48 & \textgreater{}24h & 4.2GB & \textless{}1 \\
GSDF & 26.05 & 0.46 & 3h & 1.7GB & \textless{}1 \\
3DGS & 25.01 & 0.19 &\textbf{ \textcolor{OliveGreen}{14.3m}} & \textcolor{RoyalBlue}{0.41GB} & \textbf{\textcolor{OliveGreen}{111}} \\
PGSR & \textcolor{RoyalBlue}{26.12} & \textcolor{RoyalBlue}{0.60} & \textcolor{RoyalBlue}{1.2h} & 0.42GB & \textcolor{RoyalBlue}{103} \\ \hline
CarGS & \textbf{\textcolor{OliveGreen}{26.41} }& \textbf{\textcolor{OliveGreen}{0.65}} & \textcolor{RoyalBlue}{1.2h} & \textbf{\textcolor{OliveGreen}{0.16GB}} & 90 \\ \hline
\end{tabular}
\caption{\textbf{Comparison of efficiency.} Our CarGS maintains high rendering and reconstruction quality while achieving notable computational and storage efficiency.}\label{t_eff}
\end{center}\vspace{-5mm}
\end{table}

\noindent \textbf{Efficiency Comparison.} In terms of efficiency, we compare the training time, rendering speed, and storage requirements of different methods on the “Truck” scene from the TnT dataset, as shown in Tab.~\ref{t_eff}. Our method significantly reduces storage requirements, effectively inheriting the compact characteristics of anchor-based models. Compared to the dual-model approach GSDF, our method substantially reduces training time and achieves real-time rendering. In comparison to the Gaussian primitive regularization method PGSR, our model maintains high rendering and reconstruction quality with only a small FPS decrease. Notably, our method only consumes 38\% storage than PGSR.  
\subsection{Ablation Studies}
We evaluate the effectiveness of each individual module using the `Truck' scenes from the TnT dataset. As shown in Tab.~\ref{t_abl} (a), directly applying the geometry loss (similar to the primitive-based branch) to our anchor-based method~\cite{lu2024scaffold} reveals significant limitations in reconstruction quality. (b) While employing Geo-MLP to adjust the contribution of primitives leads to noticeable improvements in reconstruction quality, overfitting to the geometry loss indicates there is still room for further enhancement. (c) By incorporating the proposed Lite-Geo, the model effectively utilizes depth information from rendering supervision, resulting in a substantial performance boost for both tasks. (d) Finally, the inclusion of geometry-guided densification enables the Gaussian primitives to capture finer details, further enhancing overall performance.

\begin{table}[]
\begin{center}
\small
\tabcolsep=0.14cm
\renewcommand{\arraystretch}{1.3}
\begin{tabular}{c|ccc|c|c}
\Xhline{1px}\hline
Methods & Geo-MLP & Lite-Geo & Geo-Densify & F1$\uparrow$ & PSNR$\uparrow$ \\ \hline
(a) &  &  &  & 0.38 & 25.72 \\
(b) & \checkmark &  &  & 0.44 & 25.75 \\
(c) & \checkmark & \checkmark &  & 0.62 & 26.13 \\
(d) & \checkmark & \checkmark & \checkmark & 0.65 & 26.41 \\ \hline
\end{tabular}
\caption{\textbf{Ablation study on the Truck scene of TnT dataset}.}\label{t_abl}
\end{center}\vspace{-5mm}
\end{table}

\section{Conclusion}
This work proposes CarGS, a unified model that leverages Contribution-adaptive Regularization to effectively address the conflicting requirements of high-quality rendering and accurate surface reconstruction. By introducing a compact MLP for adaptive contribution learning and a geometry-guided densification strategy, CarGS enhances the mutual learning between rendering and reconstruction tasks. Extensive experiments validate that CarGS achieves state-of-the-art results in both rendering fidelity and reconstruction accuracy, while maintaining real-time speed and minimal storage requirements. We believe our approach establishes a strong baseline for unified rendering and reconstruction.



{
    \small
    \bibliographystyle{ieeenat_fullname}
    \bibliography{main}
    
}


\end{document}